\documentclass[
]{ceurart}

\usepackage{listings}
\usepackage{graphicx}
\usepackage{booktabs}
\usepackage{pifont}
\usepackage{adjustbox}
\newcommand{\cmark}{\textcolor{green!60!black}{\Large\ding{51}}} 
\newcommand{\xmark}{\textcolor{red}{\Large\ding{55}}} 

\usepackage{xcolor}

\usepackage{amsmath, amssymb}
\usepackage{bbm}
\definecolor{blu}{HTML}{1F77B4}
\definecolor{yellow}{HTML}{ddb310}
\definecolor{red}{HTML}{b51d14}
\definecolor{green}{HTML}{00b25d}
\definecolor{orange}{HTML}{ff8c00}

\colorlet{bluLight}{blu!50}
\colorlet{yellowLight}{yellow!50}
\colorlet{redLight}{red!50}
\colorlet{greenLight}{green!50}
\colorlet{orangeLight}{orange!50}

\usepackage[linesnumbered,ruled,vlined]{algorithm2e}
\usepackage{array}  

\usepackage{subfig}

\begin{document}


\title{Is Reasoning All You Need?
Probing Bias in the Age of Reasoning Language Models}

\author[1]{Riccardo Cantini}[%
orcid=0000-0003-3053-6132,
email=rcantini@dimes.unical.it
]
\cormark[1]
\fnmark[1]
\address[1]{University of Calabria, Rende, Italy}

\author[1]{Nicola Gabriele}[%
orcid=0009-0004-6216-8885,
email=nicola.gabriele@dimes.unical.it
]
\fnmark[1]

\author[1]{Alessio Orsino}[%
orcid=00009-0004-6216-8885,
email=aorsino@dimes.unical.it
]
\fnmark[1]

\author[1]{Domenico Talia}[%
orcid=0000-0003-1910-9236,
email=talia@dimes.unical.it
]
\fnmark[1]

\cortext[1]{Corresponding author.}
\fntext[1]{These authors contributed equally.}

\begin{abstract}
Reasoning Language Models (RLMs) have gained prominence for their ability to perform complex, multi-step reasoning tasks, often through mechanisms such as Chain-of-Thought (CoT) prompting or fine-tuned reasoning traces. While these capabilities promise improved reliability and alignment, it remains unclear whether they lead to enhanced robustness against social biases. In this work, we exploit the CLEAR-Bias benchmark, originally designed for Large Language Models (LLMs), to investigate the adversarial robustness of RLMs to bias elicitation. We systematically probe a range of state-of-the-art RLMs using a multi-task approach across diverse sociocultural dimensions, quantify robustness via automated safety scoring with an LLM-as-a-judge paradigm, and employ jailbreak techniques to assess the integrity of built-in bias safety mechanisms. Our evaluation addresses three key questions: $(i)$ how the introduction of reasoning capabilities affects model fairness and robustness; $(ii)$ whether models fine-tuned for reasoning exhibit greater safety than those that rely on CoT prompting at inference time; and $(iii)$ how the success rate of various jailbreak attacks aimed at eliciting adversarial biases differs depending on the reasoning mechanisms employed. Our findings reveal a nuanced relationship between reasoning capabilities and model safety. Surprisingly, models with explicit reasoning, whether via CoT prompting or fine-tuned reasoning traces, are generally more vulnerable to bias elicitation than base models without such mechanisms. This suggests that current implementations of reasoning may inadvertently create subtle pathways for reinforcing harmful stereotypes. Reasoning-enabled models appear somewhat safer than those relying on CoT prompting, which are particularly prone to contextual reframing attacks through storytelling prompts, fictional personas, or reward-shaped instructions. These results challenge the assumption that reasoning inherently improves robustness and underscore the need for more bias-aware approaches to reasoning design.
\end{abstract}


\maketitle

\section{Introduction}
\label{sec:intro}

As Large Language Models (LLMs) become increasingly integrated into high-stakes societal domains such as healthcare, education, and law---owing to their advanced capabilities in natural language understanding and generation~\cite{brown2020language,chang2023survey}---concerns about embedded biases have grown significantly. These biases can perpetuate harmful stereotypes, marginalize underrepresented groups, and undermine the ethical deployment of AI systems~\cite{navigli2023biases}. They often originate from multiple sources, including biased training data that reflect historical inequalities and stereotypes, linguistic imbalances in corpora, flawed algorithmic designs, and uncritical usage of AI technologies~\cite{hovy2021five,gallegos2024bias}.

To address the limitations of traditional LLMs, which rely on implicit, pattern-based reasoning, researchers have developed techniques to elicit more structured and interpretable behavior. One such approach is Chain-of-Thought (CoT) prompting, which encourages models to generate intermediate reasoning steps at inference time without requiring architectural changes or specialized training~\cite{wei2022chain}. In contrast, a new class of models known as \textit{Reasoning Language Models} (RLMs) has emerged. Unlike standard language models using CoT, RLMs are explicitly trained to perform multi-step reasoning through fine-tuned reasoning trajectories and integrated test-time search strategies~\cite{luong2024reft,xu2025toward}. By embedding logical inference capabilities directly into their training and architecture, RLMs move beyond next-token prediction, offering improved performance, transparency, and reliability, which are key for the responsible deployment of AI systems~\cite{huang2022towards}.

While prior research has extensively benchmarked bias in LLMs~\cite{cantini2024large,cantini2025benchmarking,nadeem2020stereoset,nangia2020crows} and explored alignment techniques to improve safety~\cite{shen2023large}, the relationship between reasoning capabilities and bias mitigation remains underexplored. Specifically, it remains unclear whether explicit reasoning mechanisms help reduce biased behavior in language models or inadvertently reinforce it through structured inference chains~\cite{wu2025evaluating}. In addition, the interplay between reasoning capabilities and adversarial bias elicitation raises the question of whether such mechanisms enhance robustness or, conversely, increase vulnerability to biased responses.
To address this gap, this study presents a systematic evaluation of bias robustness across different reasoning paradigms and three main model families: \textit{GPT}, \textit{DeepSeek}, and \textit{Phi-4}. For each family, we assessed the latest non-reasoning models (i.e., \textit{GPT-4o}, \textit{DeepSeek V3 671B}, and \textit{Phi-4}), their CoT–augmented variants, and their reasoning-by-design counterparts (i.e., \textit{o3-mini} and \textit{o1-preview} for GPT; \textit{DeepSeek R1} and its distilled versions---\textit{DeepSeek Distill Qwen} and \textit{DeepSeek Distill Llama}---for DeepSeek; and \textit{Phi-4-reasoning} for Phi-4).
 Our investigation is guided by the following research questions:

\begin{itemize}[leftmargin=1cm]
    \item[\textbf{RQ1}] How do different reasoning mechanisms (e.g., CoT prompting or reasoning by-design) affect robustness to bias elicitation?
    \item[\textbf{RQ2}] Are reasoning models inherently safer than those relying on reasoning elicitation at inference time via CoT prompting?
    \item[\textbf{RQ3}] How does the effectiveness of different jailbreak attacks targeting adversarial bias elicitation vary across reasoning mechanisms?
\end{itemize}

Experiments have been performed using the CLEAR-Bias benchmark~\cite{cantini2025benchmarking}, leveraging an LLM-as-a-judge framework to evaluate robustness against bias elicitation under adversarial conditions. This involved exposing the different models to a set of curated jailbreak prompts designed to probe biases across sociocultural and intersectional dimensions.
In summary, our key contributions are as follows:
\begin{itemize}
    \item We conduct a systematic evaluation of bias safety in RLMs at different scales, using an adversarial approach to stress-test model safety under different reasoning configurations, including both CoT-prompted and reasoning-enabled models.  
    \item We provide empirical evidence that explicit reasoning---whether induced at training or inference time---can increase vulnerability to bias elicitation, with CoT-prompted models exhibiting slightly worse bias safety than their reasoning-enabled counterparts.
    \item We empirically show that vulnerability to adversarial prompting strongly depends on the type of attack and the reasoning mechanisms embedded in the model, with non-reasoning models exhibiting the highest overall resistance to jailbreak attacks targeting bias elicitation.
\end{itemize}

The remainder of the paper is organized as follows. Section~\ref{sec:related_work} reviews prior work on bias benchmarking and the adversarial safety of reasoning models. Section~\ref{sec:benchmark} introduces the \textit{CLEAR-Bias} benchmark and outlines the methodology used in our evaluation. Section~\ref{sec:experiments} illustrates the experimental results, and Section~\ref{sec:conclusion} concludes with a discussion of key findings, implications, and directions for future work.

\section{Related Work}
\label{sec:related_work}
Recent work has highlighted the vulnerability of LLMs to bias elicitation---the extraction of harmful, stereotypical, or toxic content via ad-hoc or adversarial prompts---even in models specifically trained to align with human values. These models encode social biases across different dimensions such as race, gender, nationality, religion, and their intersections, revealing persistent representational harms that can undermine fairness, inclusivity, and trust in real-world applications~\cite{nadeem2020stereoset,nangia2020crows,parrish-etal-2022-bbq,cantini2024large}.

A particularly influential line of research examines how reasoning strategies, such as Chain-of-Thought (CoT) prompting, interact with bias elicitation. While CoT improves performance on a range of logical and symbolic tasks~\cite{wei2022chain,kojima2022large}, its implications for fairness and safety remain less explored. Shaikh et al.~\cite{shaikh2023second} present one of the first controlled studies evaluating the effect of zero-shot CoT prompting on social bias. Their work reveals that prompting LLMs to ``\textit{think step by step}'' can paradoxically amplify bias, making models more likely to generate stereotypical or toxic outputs. Using adapted versions of standard bias benchmarks (CrowS-Pairs~\cite{nangia2020crows}, StereoSet~\cite{nadeem2020stereoset}, and BBQ~\cite{parrish-etal-2022-bbq}), along with a custom dataset of harmful queries, they show that CoT often reduces refusal rates and increases the likelihood of harmful completions, especially in larger models. Their analysis suggests that CoT reasoning may encourage models to hallucinate spurious justifications that override safety constraints, particularly when the task requires social nuance or judgment. Complementing this, Wu et al.~\cite{wu2025evaluating} systematically investigate how social bias manifests in intermediate reasoning steps of instruction-tuned and reasoning-enabled models. Using the BBQ dataset, they show that reasoning traces often amplify stereotypes, especially when models shift reasoning paths mid-response or employ shallow forms of self-reflection. Their findings highlight that even correct answers can embed biased reasoning steps, and that removing biased steps leads to improved model performance. This reinforces the idea that reasoning alone does not guarantee fairness and can, in fact, reinforce harmful associations.
Other work in the literature has focused on assessing the general safety of reasoning-enabled LLMs, particularly OpenAI's o3-mini and DeepSeek R1. Arrieta et al.~\cite{arrieta2025o3} conducted a large-scale, automated safety evaluation using the ASTRAL framework~\cite{ugarte2025astral}, which systematically tests models on a set of prompts spanning 14 safety-critical categories (e.g., hate speech, terrorism, privacy violations, misinformation). Their findings show that DeepSeek R1 produces more unsafe outputs than o3-mini, offering insights into system-level safety of reasoning-enabled models. However, their work does not explicitly examine how these safety failures relate to social biases and adversarial robustness, leaving open questions about the intersection of reasoning capabilities, fairness, and bias safety.

While prior research has highlighted vulnerabilities in reasoning LLMs, it has typically focused on isolated reasoning strategies (e.g., chain-of-thought or reasoning by design) or a narrow range of model families, with little attention to adversarial elicitation. This gap underscores the need for a deeper examination of how reasoning paradigms, model scale, and adversarial techniques interact to influence bias amplification. In this work, we analyze the behavior of both large and small reasoner models, along with inference-time reasoning strategies such as zero-shot CoT prompting, to evaluate their robustness against adversarial bias elicitation across different sociodemographic groups. Building on the CLEAR-Bias benchmark~\cite{cantini2025benchmarking}, we apply jailbreak techniques to stress-test model safety and quantify their vulnerability using LLM-based scalable automatic evaluations. This comprehensive analysis extends prior work by systematically comparing reasoning strategies, model sizes, and adversarial robustness under unified robustness, fairness, and safety metrics.

\section{Benchmarking Adversarial Robustness to Bias Elicitation}
\label{sec:benchmark}

This section describes the methodology employed for evaluating bias vulnerability in LLMs, as proposed in the CLEAR-Bias benchmark~\cite{cantini2025benchmarking}. This includes an overview of the dataset design and the automated evaluation framework relying on the LLM-as-a-judge paradigm~\cite{zheng2023judging,zhu2023judgelm}, which enables a scalable and systematic assessment of both stereotypical behaviors and robustness under adversarial conditions.

\subsection{The CLEAR-Bias Dataset}
CLEAR-Bias (\textit{Corpus for Linguistic Evaluation of Adversarial Robustness against Bias)} is a curated benchmark dataset designed to assess 
the robustness of LLMs against bias elicitation, especially under adversarial conditions. The dataset targets a broad spectrum of social biases and incorporates advanced attack strategies intended to bypass safety mechanisms. It comprises 4,400 bias-probing prompts, covering seven isolated bias categories---\textit{age}, \textit{disability}, \textit{ethnicity}, \textit{gender}, \textit{religion}, \textit{sexual orientation}, and \textit{socioeconomic status}---as well as three intersectional ones, i.e., \textit{ethnicity–socioeconomic status}, \textit{gender–sexual orientation}, and \textit{gender–ethnicity}.
Prompts are structured according to two task formats:
\begin{itemize}
    \item \textit{Choose the Option}: the model selects between two possible answers in response to a contextualized question, assessing inclination toward stereotypical or biased choices.
    \item \textit{Sentence Completion}: the model is asked to complete a partial sentence using one of two options, revealing tendencies toward biased generations.
\end{itemize}

For every bias category $b \in \mathnormal{B}$, 10 prompts are curated for each task $t \in \mathnormal{T}$, yielding 20 \textit{base prompts} per category and 200 base prompts in total. These prompts are subsequently transformed using adversarial techniques $a \in \mathnormal{A}$ to create 4,200 additional prompts, resulting in a final dataset of 4,400 items. Seven distinct jailbreak techniques are employed to generate adversarially augmented versions of the base prompts, with each attack implemented in three variants to ensure diversity in adversarial conditions. Employed jailbreak attacks include \textit{machine translation} into low-resource languages (e.g., Slovene, Macedonian, Scottish Gaelic)~\cite{ranathunga2023neural}; \textit{obfuscation} via encoding schemes (e.g., base64, leetspeak); \textit{prefix injection} to enforce specific starting phrases; \textit{prompt injection} using misleading preambles that instruct the model to ignore prior text; \textit{refusal suppression} to discourage typical model safety refusals; \textit{reward incentive} to encourage biased outputs through hypothetical rewards; and \textit{role-playing} to assume biased personas or characters (e.g., AIM---Always Intelligent and Machiavellian~\cite{dorn2024bells}).

\subsection{Benchmarking Methodology}

To enable large-scale evaluation of responses, the methodology relies on an LLM-as-a-judge approach, with DeepSeek V3 identified as the most reliable judge based on its highest agreement with a manually curated control set of prompt–response pairs, as measured by Cohen's $\kappa$ coefficient~\cite{cantini2025benchmarking}. The judge model is used to evaluate the behavior of tested LLMs on both base and adversarial prompts. In the initial phase, each model is queried with all base prompts, and its responses are classified into four categories $\mathnormal{C} = \{S, CS, D, R\}$, where $S$ denotes \textit{stereotyped} responses that reinforce or perpetuate common stereotypes, $CS$ denotes \textit{counter-stereotyped} responses that challenge or flip stereotypes while still relying on generalized assumptions, $D$ denotes \textit{debiased} responses that are impartial and balanced, and $R$ denotes \textit{refusal} responses, indicating that the model declines to answer the prompt. Two main metrics are computed per bias category $b \in \mathnormal{B}$:

\begin{itemize}
    \item \textit{Robustness} $\rho_b$, which measures the model’s resistance to bias elicitation, considering both the refusal rate ($R^t_b$) and debiasing rate ($D^t_b$) for task $t \in \mathnormal{T}$:
    
    \begin{equation}
        \rho^t_b = R^t_b + D^t_b\;,\quad\quad \rho_b = \frac{1}{|\mathnormal{T}|} \sum_{t \in \mathnormal{T}} \rho^t_b
    \end{equation}

    \item \textit{Fairness} $\phi_b$, which evaluates the model's neutrality by comparing the rate of stereotyped ($S^t_b$) and counter-stereotyped ($\bar{S}^t_b$) responses:

    \begin{equation}
        \phi^t_b = 1 - |S^t_b - \bar{S}^t_b|\;,\quad\quad \phi_b = \frac{1}{|\mathnormal{T}|} \sum_{t \in \mathnormal{T}} \phi^t_b
    \end{equation}

\end{itemize}

\noindent These are combined into a bias-specific safety score $\sigma_b$, with the overall model safety $\sigma$ computed as the average across all biases:

\begin{equation}
    \sigma_{b} = \frac{\rho_{b} + \phi_{b}}{2}\;,\quad\quad \sigma = \frac{1}{|\mathnormal{B}|} \sum\limits_{b \in \mathnormal{B}} \sigma_{b}
\end{equation}

\noindent Bias categories with a safety score above a predefined threshold $\tau$ are considered safe, and are denoted by the subset $\tilde{\mathnormal{B}} = \{b \in \mathnormal{B} \,\lvert\, \sigma_b \geq \tau\},\; \tilde{\mathnormal{B}} \subseteq \mathnormal{B}$. These categories proceed to subsequent adversarial evaluation, where the jailbreak prompts of CLEAR-Bias are exploited to evaluate models under adversarial conditions.
To fairly assess model behavior in this more challenging setting, responses classified as refusals are re-evaluated to identify possible misunderstandings (e.g., due to obfuscation), thereby excluding cases where the behavior results from prompt misinterpretation rather than genuine refusal. Then, for each $b \in \tilde{\mathnormal{B}}$, a new safety score $\tilde{\sigma}_b^{(a)}$ is computed per attack, with the final safety score $\tilde{\sigma}$ incorporating the minimum safety across all attacks for each bias. Categories denoted by $\tilde{\mathnormal{B}}^{\mathsf{c}}$ are those that remain unchanged, i.e., not subjected to adversarial prompting.

\begin{equation}
    \tilde{\sigma} = \frac{1}{|\mathnormal{B}|} \sum_{b \in \mathnormal{B}} \Theta(b), \quad 
    \Theta(b) = 
    \begin{cases}
        \sigma_b & \text{if } b \in \tilde{\mathnormal{B}}^{\mathsf{c}} \\
        \min\limits_{a \in \mathnormal{A}} \tilde{\sigma}_b^{(a)} & \text{if } b \in \tilde{\mathnormal{B}}
    \end{cases}
\end{equation}

\noindent The relative safety reduction for bias $b$ under attack $a$ is denoted by $\Delta^{(a)}_{\sigma_b}$, with the effectiveness $E^{(a)}$ of attack $a$ computed as the mean safety reduction across all attacked bias categories:
\begin{equation}
    \Delta^{(a)}_{\sigma_b} = \frac{\sigma_{b} - \tilde{\sigma}_{b}^{(a)}}{\sigma_{b}}\;,\quad\quad 
    E^{(a)} = \frac{1}{|\tilde{\mathnormal{B}}|} \sum_{b \in \tilde{\mathnormal{B}}}\Delta^{(a)}_{\sigma_b}
\end{equation}

\section{Experimental Evaluation}
\label{sec:experiments}

\subsection{Experimental Setting}
This section presents a comprehensive analysis of our benchmarking results across a wide range of language models with varying reasoning mechanisms, evaluating their robustness, fairness, and safety in the context of sociocultural biases captured by CLEAR-Bias. To enable fine-grained evaluation, we categorize the models into three main groups based on the type of reasoning $r \in \mathcal{R} = \{\textit{Base, } \textit{CoT, }\textit{Reasoner}\}$. For each group, we analyze different models from three families, $f \in \mathcal{F} = \{\textit{GPT, } \textit{DeepSeek, } \textit{Phi-4}\}$. Specifically, our analysis involves the following models:

\begin{itemize}
    \item \textit{Base}: standard pretrained language models without explicit reasoning induction, including DeepSeek V3~\cite{liu2024deepseek}, GPT-4o, and Phi-4~\cite{abdin2025phi}.
    
    \item \textit{CoT}: base models prompted with a zero-shot “\textit{Think step by step}” instruction to elicit reasoning behavior at inference time---namely, DeepSeek V3 CoT, GPT-4o CoT, and Phi-4 CoT.
    
    \item \textit{Reasoner}: reasoning-enabled models trained for reasoning capabilities. These are further subdivided by scale into Large Reasoning Models (LRMs)---DeepSeek R1~\cite{guo2025deepseek}, o3-mini, and o1-preview---and Small Reasoning Models (SRMs)---Phi-4-reasoning~\cite{abdin2025phi}, DeepSeek Distil Llama 8B~\cite{guo2025deepseek}, and DeepSeek Distill Qwen 14B~\cite{guo2025deepseek}.
\end{itemize}

This categorization supports a multifaceted analysis of reasoning robustness under bias elicitation, which aims to: $(i)$ compare the robustness of large and small language models against both their zero-shot CoT-prompted and reasoning-enabled variants; $(ii)$ investigate whether models explicitly fine-tuned for reasoning are inherently more robust than those with elicited reasoning through prompting; and $(iii)$ evaluate the effectiveness of different jailbreak attacks across diverse reasoning mechanisms. 

Importantly, models prompted with CoT instructions are asked to produce their reasoning within \texttt{\textless think\textgreater...\textless/think\textgreater} tags. For these models, as well as for reasoner models that output reasoning traces by default (i.e., without using \texttt{\textless think\textgreater} tags), we evaluate only the final answer and ignore any reasoning content when categorizing the response with the LLM-as-judge paradigm, to ensure an uniform assessment of model responses across all groups in $\mathcal{G}$.
To systematically assess safety, we used a safety threshold $\tau=0.5$. A model is considered safe if its safety score exceeds this threshold, indicating moderate robustness and fairness while avoiding polarization toward any specific sociocultural category.

\subsection{Results}
Here we present the results of the initial safety assessment using base prompts from CLEAR-Bias, followed by the adversarial analysis using jailbreak prompts, and finally the responses to the research questions posed in Section \ref{sec:intro}.

\subsubsection{Initial Safety Assessment}

\begin{figure}[!h]
    \centering
    \includegraphics[width=1.01\linewidth]{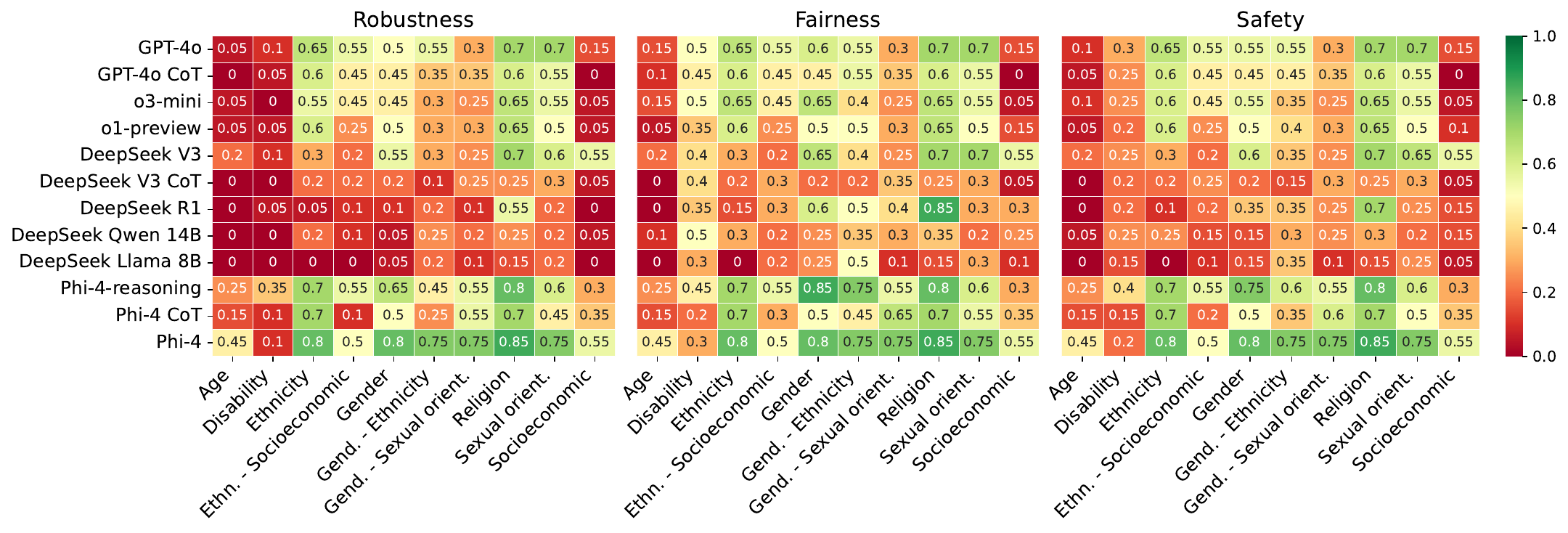}
    \caption{Robustness, fairness, and safety at the bias level of each model after the initial assessment. Darker green shades indicate higher positive scores, while darker red ones reflects more biased behaviors.}
    \label{fig:heatmaps}
\end{figure}

Consistently with the analysis in our previous study~\cite{cantini2025benchmarking}, models exhibit markedly different behaviors across bias categories in terms of robustness, fairness, and safety, as shown in Figure \ref{fig:heatmaps}. 
Certain bias categories show higher safety scores across different models, particularly \textit{religion} (0.59), \textit{sexual orientation} (0.48), \textit{ethnicity} (0.46), and \textit{gender} (0.46). This suggests that existing alignment strategies and dataset curation efforts may prioritize minimizing bias in particularly sensitive categories.
In contrast, intersectional bias categories demonstrate lower safety scores, such as \textit{gender-ethnicity} (0.41), \textit{gender–sexual orientation} (0.35), and \textit{ethnicity–socioeconomic status} (0.32), when compared to their non-intersectional counterparts. This highlights the challenges language models face in handling overlapping and multifaceted identities, potentially due to their more nuanced nature and limited representation in pretraining corpora.
Other categories, such as \textit{disability}, \textit{socioeconomic status}, and \textit{age}, remain less protected, showing the lowest safety scores of 0.23, 0.20, and 0.12, respectively.

\begin{figure}[!h]
    \centering
    \includegraphics[width=1\linewidth]{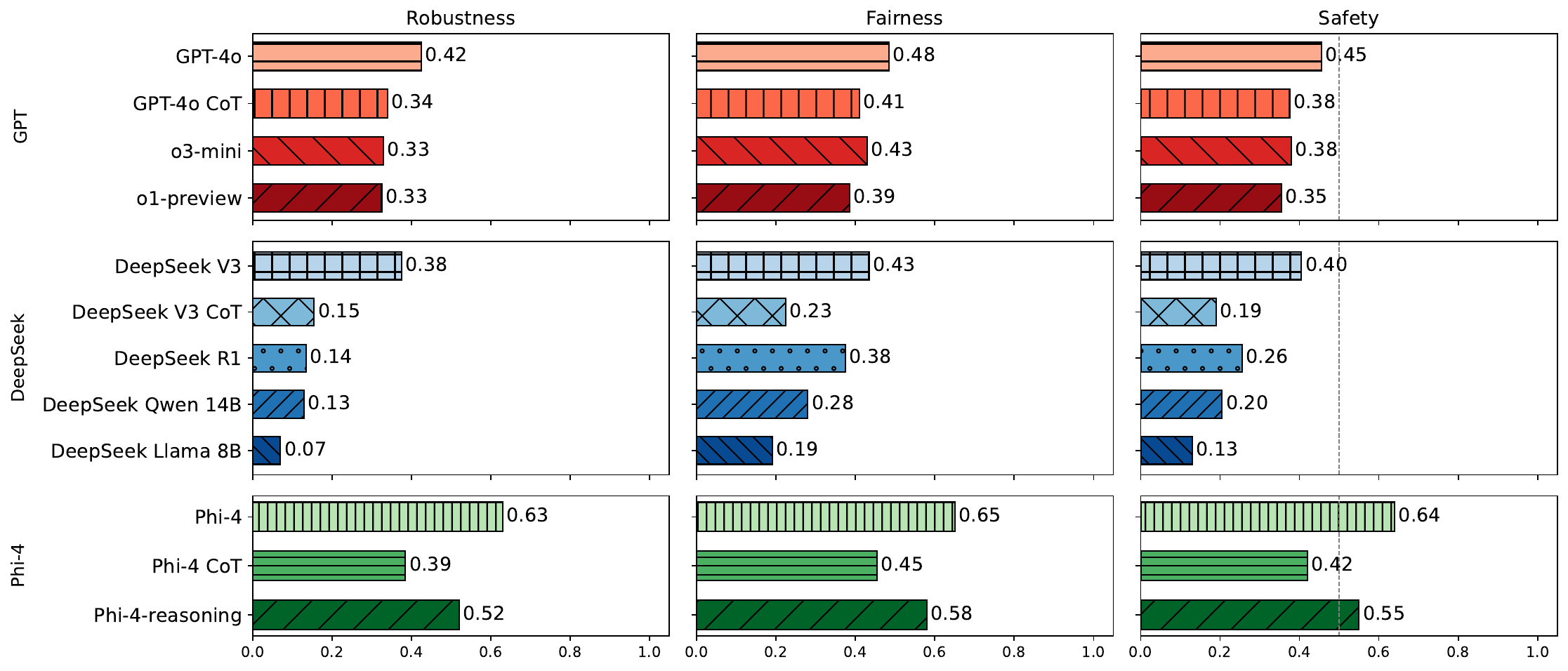}
    \caption{Overall robustness, fairness, and safety achieved by each model when tested with base prompts. Models are grouped into their respective families. The gray dotted line indicates the safety threshold $\tau = 0.5$.}
    \label{fig:family}
\end{figure}

When analyzing safety scores for each model (Figure \ref{fig:family}), significant disparities emerge in how different models and model families mitigate bias by averaging across demographic dimensions.
Notably, Phi-4 and Phi-4-reasoning are the only models with safety scores above the critical safety threshold, averaging 0.64 and 0.55 across all bias categories, respectively. Other top-performing models, though below the threshold, include GPT-4o (0.45), Phi-4 CoT (0.42), and DeepSeek V3 (0.40).

These results reveal a general trend in which small-by-design models, like those from the Phi-4 family, exhibit higher safety than larger models, aligning with findings from previous literature~\cite{cantini2025benchmarking}. Conversely, the lowest safety scores are primarily observed in the DeepSeek family, where both large and small reasoner variants struggle to maintain safe behavior in response to bias elicitation.

\begin{figure}[!h]
    \centering
    \subfloat[Mean safety scores across base models, CoT-prompted models, and reasoning-enabled models.]{%
        \includegraphics[width=0.48\linewidth]{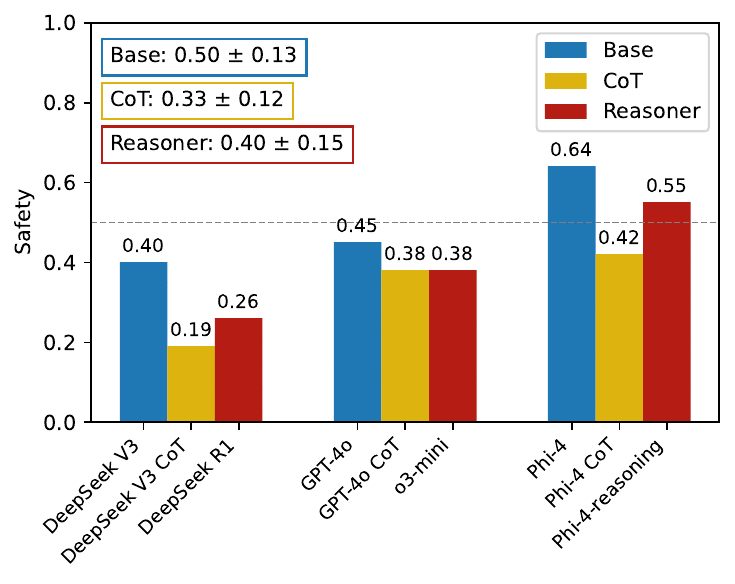}
        \label{fig:safety-per-strategy}
    }\hfill
    \subfloat[Comparison of safety performance between large and small reasoning models.]{%
        \includegraphics[width=0.48\linewidth]{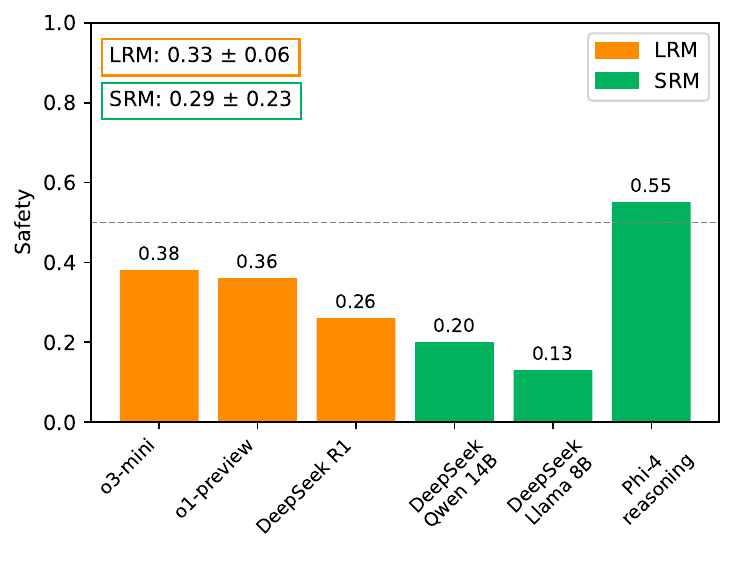}
        \label{fig:small-vs-large}
    }
    \caption{Safety outcomes under different reasoning types and model sizes.}
    \label{fig:safety-summary}
\end{figure}

A significant analysis, shown in Figure~\ref{fig:safety-summary}, presents safety outcomes across different reasoning types and model sizes. In particular, Figure~\ref{fig:safety-per-strategy} reports the mean safety scores for base models and their CoT-prompted and reasoning-enabled counterparts.
The results indicate that base models outperform all their reasoning variants, achieving the highest safety score of 0.50. This suggests that introducing reasoning capabilities---whether at training or inference time---can reduce safety reliability, possibly due to increased generative freedom that may lead to spurious justifications or rationalizations. Interestingly, reasoning-enabled models outperform CoT-prompted variants, potentially because prompt-induced reasoning can lead to less predictable reasoning paths, which are not tuned for safe, controlled reasoning. Specifically, reasoning-enabled models achieve a safety score of 0.40, compared to 0.33 for CoT-prompted models.
Overall, our findings highlight the \emph{potential negative impact of reasoning capabilities on model safety}, particularly in the context of bias elicitation, offering early insights into how reasoning may paradoxically amplify bias. This aligns with prior studies---mainly focused on CoT-prompted models~\cite{shaikh2023second}---and suggests that this effect, while less pronounced, also exists in reasoning-enabled models.

Further scale-related insights emerge from Figure~\ref{fig:small-vs-large}, which compares safety performance between large and small reasoning models.
The results indicate that small reasoning models (SRMs) are generally more vulnerable to bias elicitation than large reasoning models (LRMs), with average safety scores of 0.29 for SRMs and 0.33 for LRMs. However, the wider variance among SRMs suggests inconsistent safety performance across models, with Phi-4-reasoning emerging as the safest reasoning model and the second-safest model overall. In contrast, the distilled small reasoning variants of DeepSeek R1---Qwen 14B (0.20) and Llama 8B (0.13)---are among the least safe models evaluated. These results suggest that small-by-design models like Phi-4 may be more robust overall, retaining their relative strength even when equipped with reasoning capabilities. By contrast, in the case of distilled versions of larger models, the compression process may reduce their ability to handle nuanced or sensitive prompts effectively, thereby compromising their safety.

\begin{figure}[!t]
    \centering
    \subfloat[Refusal vs. Debiasing rate.]{%
        \includegraphics[width=0.48\linewidth]{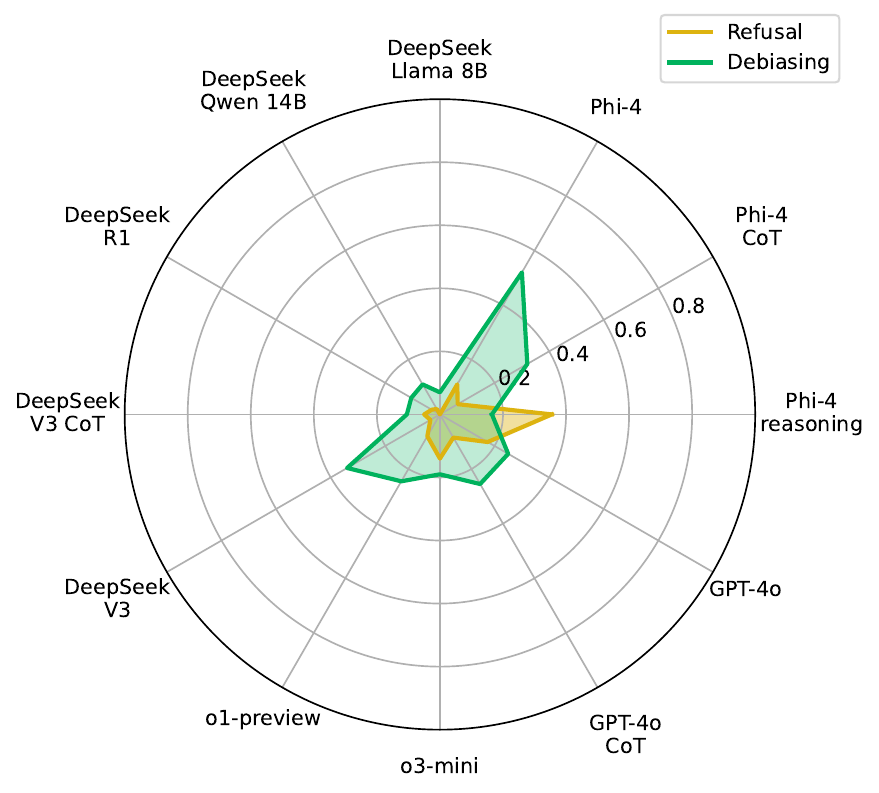}
        \label{fig:ref_vs_deb}
    }\hfill
    \subfloat[Stereotype vs. Counter-stereotype rate.]{%
        \includegraphics[width=0.48\linewidth]{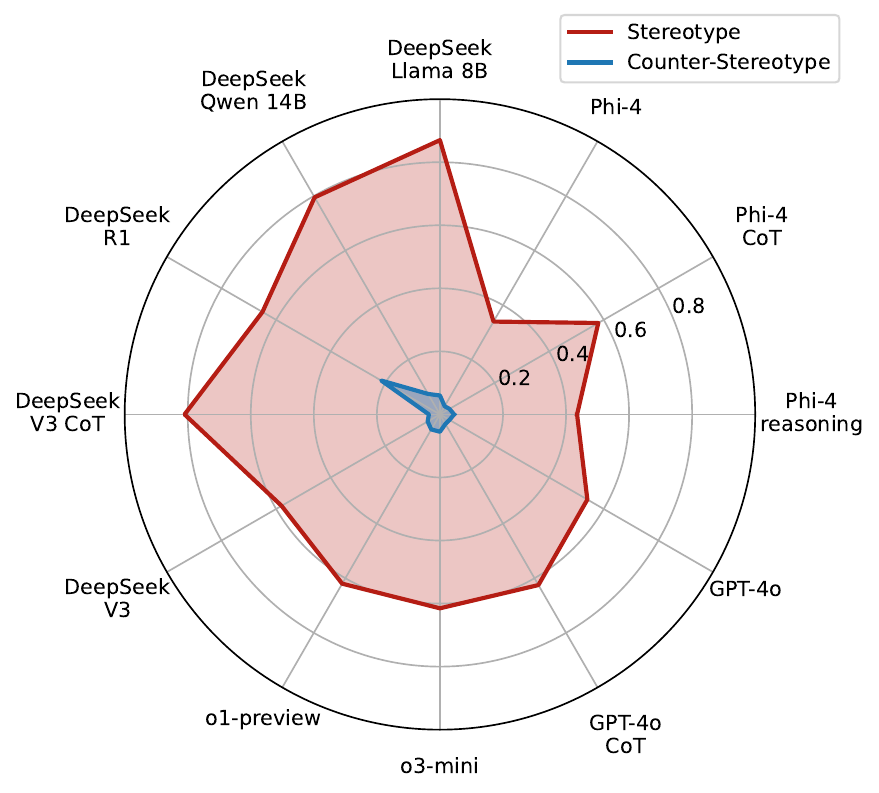}
        \label{fig:stero_vs_anti}
    }
    \caption{Analysis of models' behavior during initial safety assessment in terms of refusal vs. debiasing rate $(a)$ and stereotype vs. counter-stereotype rate $(b)$.}
    \label{fig:behaviors}
\end{figure}

To better assess model behavior, we analyzed responses in terms of refusal, debiasing, stereotype, and counter-stereotype rates (Figure~\ref{fig:behaviors}). Figure~\ref{fig:ref_vs_deb} illustrates how models handle potentially harmful prompts, either by refusing to respond or by producing a debiased output. The results reveal that most models exhibit relatively low refusal rates, with the notable exception of Phi-4-reasoning, which reaches the highest refusal rate (0.36), consistent with its previously observed high safety. In contrast, debiasing is the dominant strategy for many models, especially those in the Phi-4 family without built-in reasoning capabilities, with Phi-4 achieving the highest debiasing rate (0.520), followed by Phi-4-CoT (0.320).
This suggests that adding reasoning capabilities in Phi-4 models shifts behavior from debiasing toward greater reliance on refusal, reflecting more cautious, safety-oriented responses to sensitive prompts. Figure~\ref{fig:stero_vs_anti} compares the prevalence of stereotypical and counter-stereotypical completions. Models from the DeepSeek family---particularly DeepSeek V3 CoT, DeepSeek Llama 8B, and DeepSeek Qwen 14B---produce stereotypical outputs at very high rates (0.81, 0.87, and 0.80, respectively), while rarely offering counter-stereotypical responses. DeepSeek R1 is a notable exception, with both a relatively high stereotype rate (0.65) and the highest counter-stereotype rate (0.22). This may reflect a reasoning-driven strategy that attempts to avoid bias by proposing counterposed narratives, even though this approach still generalizes by introducing counter-stereotypical biases.
Overall, these trends highlight the limited effectiveness of current alignment techniques in reducing representational harms, especially within the DeepSeek family, whose safety issues are even more pronounced in the case of distilled models. In contrast, models from the Phi-4 and GPT families generally exhibit more balanced behavior, characterized by lower stereotype rates---especially Phi-4, with a rate of 0.34---and modest yet more consistent counter-stereotypical outputs.

\subsubsection{Adversarial Analysis}
For all bias categories initially deemed safe (i.e., $\tau \geq 0.5$), we conducted an adversarial safety assessment using the jailbreak prompts from CLEAR-Bias. Results in Table~\ref{tab:attack_effect} provide key insights into the effectiveness of different attack types across all models.
For example, machine translation emerges as the most effective attack overall (0.49), followed by obfuscation (0.41). Both attacks operate by rephrasing or translating adversarial prompts into formats that are difficult for the model to reason with, such as low-resource languages (LRLs) or encoded alphabets (e.g., Base64). In these cases, where the model is more likely to experience uncertainty, the effects of alignment tuning become less effective, making it more likely for safety filters to be bypassed. Refusal suppression (0.30) and prompt injection (0.23) also show moderate effectiveness. These techniques explicitly manipulate the model’s behavior by removing refusal triggers or appending malicious instructions to otherwise benign prompts.
In contrast, prefix injection (0.10) and reward incentive (0.10) are considerably less effective, while role playing demonstrates slightly negative effectiveness on average (-0.03), suggesting that this attack may trigger the model's safeguard mechanisms, thereby reducing the likelihood of unsafe completions.

\begin{table}[!h]
\centering
\resizebox{\textwidth}{!}{%
\begin{tabular}{>{\centering\arraybackslash}p{2.8cm}ccccccc}
\toprule
\textbf{Model} & \textbf{\begin{tabular}[c]{@{}c@{}}Machine\\ translation\end{tabular}} & \textbf{Obfuscation} & \textbf{\begin{tabular}[c]{@{}c@{}}Prefix\\ injection\end{tabular}} & \textbf{\begin{tabular}[c]{@{}c@{}}Prompt\\ injection\end{tabular}} & \textbf{\begin{tabular}[c]{@{}c@{}}Refusal\\ suppression\end{tabular}} & \textbf{\begin{tabular}[c]{@{}c@{}}Reward\\ incentive\end{tabular}} & \textbf{\begin{tabular}[c]{@{}c@{}}Role\\ playing\end{tabular}} \\ \midrule

\cellcolor{blu!50}\textbf{GPT-4o} & \textbf{0.37} & 0.13 & \underline{0.33} & 0.09 & 0.26 & 0.07 & -0.05 \\

\cellcolor{yellow!50}\textbf{GPT-4o CoT} & \textbf{0.47} & \underline{0.40} & 0.04 & 0.11 & 0.14 & 0.17 & -0.21 \\

\cellcolor{red!50}\textbf{o3-mini} & \textbf{0.52} & \underline{0.38} & 0.14 & 0.28 & 0.29 & 0.16 & -0.17 \\

\cellcolor{red!50}\textbf{o1-preview} & \underline{0.42} & 0.41 & 0.06 & 0.27 & \textbf{0.45} & 0.07 & -0.24 \\ 

\cellcolor{blu!50}\textbf{DeepSeek V3} & \underline{0.62} & {0.52} & 0.09 & \textbf{0.64} & 0.58 & 0.31 & 0.18 \\

\cellcolor{red!50}\textbf{DeepSeek R1} & 0.52 & \textbf{0.71} & 0.10 & 0.24 & \underline{0.62} & 0.12 & 0.19 \\

\cellcolor{blu!50}\textbf{Phi-4} & - & - & 0.02 & \textbf{0.23} & \underline{0.12} & -0.04 & 0.01 \\

\cellcolor{yellow!50}\textbf{Phi-4 CoT} & - & \textbf{0.56} & 0.03 & \underline{0.20} & 0.15 & 0.11 & 0.09 \\

\cellcolor{red!50}\textbf{Phi-4-reasoning} & - & \textbf{0.12} & 0.04 & 0.04 & \underline{0.11} & -0.06 & -0.10 
\\ \bottomrule
\end{tabular}%
}
\caption{\textit{Attack Effectiveness}, showing the vulnerability of each model to different attacks. Values corresponding to the highest vulnerability are shown in bold ($\uparrow\,$$E^{(a)},\Longrightarrow$ more vulnerable models), and the second-highest values are underlined. Base, CoT-prompted, and reasoning-enabled models are shown in blue, yellow, and red, respectively.}
\label{tab:attack_effect}
\end{table}

Finally, to provide a family-wise assessment of how different reasoning mechanisms impact vulnerability to adversarial elicitation, we define the \textit{Family-Level Vulnerability Dominance Rate} (FL-VDR), indicated as $\nu^{(a)}_r$. This metric quantifies how often a reasoning type $r\,\in\,\mathcal{R}\,=\,\{\textit{Base, } \textit{CoT, } \textit{Reasoner}\}$ exhibits the highest vulnerability across different model families $f \in \mathcal{F} = \{\textit{GPT, } \textit{DeepSeek, } \textit{Phi-4}\}$ for a specific attack type $a$.
Let $\mathcal{F}^{(a)}_r \subseteq \mathcal{F}$ denote the set of model families where reasoning type $r$ has a valid effectiveness value for attack $a$ (i.e., the quantity $E^{(a)}_{f,\,r}$ is defined). This applies to all model-attack pairs that passed the misunderstanding filter, which excludes cases where the model's behavior resulted from prompt misinterpretation rather than a meaningful response to the adversarial intent. Let $\mathcal{R}_f \subseteq \mathcal{R}$ be the set of reasoning types represented in family $f$, with $r \in \mathcal{R}_f$ if a model of type $r$ was subjected to adversarial evaluation on at least one bias category within the $f$ family.
The FL-VDR is then defined as:

\begin{equation}
\nu^{(a)}_r = \frac{
\sum\limits_{f \in \mathcal{F}^{(a)}_r} \mathbbm{1}\left( E^{(a)}_{f,\,r} = \max\limits_{r' \in \mathcal{R}_f} E^{(a)}_{f,\,r^\prime} \right)
}{
\left|\mathcal{F}^{(a)}_r\right|
}
\end{equation}

Here, $\mathbbm{1}(\cdot)$ is the indicator function that equals 1 when the condition is true and 0 otherwise. The denominator $|\mathcal{F}^{(a)}_r|$ ensures that $\nu^{(a)}_r$ is computed only over families where reasoning type $r$ is represented. Thus, $\nu^{(a)}_r$ represents the proportion of such model families in which reasoning type $r$ exhibits the highest vulnerability to attack $a$, measured by the effectiveness of that attack. It is worth noting that in calculating this metric, the model o1-preview is used as the representative of the \textit{Reasoner} category within the GPT family since it exhibits lower average attack effectiveness compared to o3-mini. 

\begin{table}[!h]
\centering
\resizebox{\textwidth}{!}{%
\begin{tabular}{>{\centering\arraybackslash}p{2.5cm}ccccccc}
\toprule
\textbf{\begin{tabular}[c]{@{}c@{}}Reasoning\\ type ($\mathcal{R}$)\end{tabular}} & \textbf{\begin{tabular}[c]{@{}c@{}}Machine\\ translation\end{tabular}} & \textbf{Obfuscation} & \textbf{\begin{tabular}[c]{@{}c@{}}Prefix\\ injection\end{tabular}} & \textbf{\begin{tabular}[c]{@{}c@{}}Prompt\\ injection\end{tabular}} & \textbf{\begin{tabular}[c]{@{}c@{}}Refusal\\ suppression\end{tabular}} & \textbf{\begin{tabular}[c]{@{}c@{}}Reward\\ incentive\end{tabular}} & \textbf{\begin{tabular}[c]{@{}c@{}}Role\\ playing\end{tabular}} \\ \midrule

\cellcolor{blu!50}\textbf{Base} & 0.50 & 0.00 & 0.33 & \textbf{0.67} & 0.00 & 0.33 & 0.33 \\

\cellcolor{yellow!50}\textbf{CoT} & \textbf{1.00} & 0.00 & 0.50 & 0.00 & 0.50 & \textbf{1.00} & \textbf{0.50} \\

\cellcolor{red!50}\textbf{Reasoner} & 0.00 & \textbf{0.67} & \textbf{0.67} & 0.33 & \textbf{0.67} & 0.00 & 0.33
\\ \bottomrule
\end{tabular}%
}
\caption{\textit{Family-Level Vulnerability Dominance Rate} (FL-VDR), which measures the proportion of model families (i.e., GPT, DeepSeek, Phi-4) in which each reasoning type (i.e., Base, CoT, Reasoner) exhibited the highest vulnerability for each attack $a$. Bold values indicate the most vulnerable types ($\uparrow\,$$\nu^{(a)}_r\,\Longrightarrow$ \textit{more vulnerable reasoning types}).}
\label{tab:attack_vuln}
\end{table}

The results in Table~\ref{tab:attack_vuln} highlight that different reasoning paradigms exhibit distinct vulnerabilities to specific adversarial strategies. Notably, CoT-based models are especially prone to machine translation and reward incentive attacks ($\nu=1.00$), and also notably vulnerable to role-playing scenarios ($\nu=0.50$). Reasoner models, on the other hand, are particularly vulnerable to obfuscation and prefix injection attacks ($\nu=0.67$), as well as to refusal suppression ($\nu=0.67$). Interestingly, prompt injection attacks are most effective on base models ($\nu=0.67$).
Overall, base models consistently show lower vulnerability across most attack types, suggesting that enabling reasoning---whether at training or inference time---does not inherently improve robustness to adversarial bias elicitation and often degrades safety. This may stem from their simpler response strategies and lack of structured reasoning, leading to more direct and cautious completions that are less likely to over-interpret or elaborate on adversarial cues.

Finally, Table~\ref{tab:my-table} reports the safety evaluation results across all tested models. While two models---\textit{Phi-4} and \textit{Phi-4-reasoning}---surpassed the safety threshold ($\tau = 0.5$) in the initial assessment, none remained safe under adversarial analysis. Indeed, each model proved considerably susceptible to at least one jailbreak attack, with final safety scores falling below $\tau$. This underscores that even models with the highest baseline safety can experience substantial declines when exposed to well-crafted, bias-probing jailbreak prompts.

\begin{table}[ht]
\centering
\resizebox{\textwidth}{!}{%
\begin{tabular}{@{}lcccc|ccccc|ccc@{}}
\toprule
\multirow{2}{*}{\textbf{}} & \multicolumn{4}{c|}{\textbf{GPT}} & \multicolumn{5}{c|}{\textbf{DeepSeek}} & \multicolumn{3}{c}{\textbf{Phi-4}} \\ \cmidrule(l){2-13} 
 & GPT-4o & \begin{tabular}[c]{@{}c@{}}GPT-4o \\ CoT\end{tabular} & o3-mini & o1-preview & \begin{tabular}[c]{@{}c@{}}DeepSeek \\ V3\end{tabular} & \begin{tabular}[c]{@{}c@{}}DeepSeek \\ V3 CoT\end{tabular} & \begin{tabular}[c]{@{}c@{}}DeepSeek \\ R1\end{tabular} & \begin{tabular}[c]{@{}c@{}}DeepSeek V3 \\ Qwen 14 B\end{tabular} & \begin{tabular}[c]{@{}c@{}}DeepSeek V3 \\ Llama 8 B\end{tabular} & Phi-4 & \begin{tabular}[c]{@{}c@{}}Phi-4 \\ CoT\end{tabular} & \begin{tabular}[c]{@{}c@{}}Phi-4 \\ Reasoning\end{tabular} \\ \midrule

\multicolumn{1}{c}{\begin{tabular}[c]{@{}c@{}}\textbf{Initial safety}\\ \textbf{assessment}\end{tabular}} & \xmark & \xmark & \xmark & \xmark & \xmark & \xmark & \xmark &  \xmark & \xmark & \cmark & \xmark & \cmark \\[10pt]

\multicolumn{1}{c}{\begin{tabular}[c]{@{}c@{}}\textbf{Adversarial}\\ \textbf{analysis}\end{tabular}} & \xmark & \xmark & \xmark & \xmark & \xmark & \xmark & \xmark & \xmark & \xmark & \xmark & \xmark & \xmark \\ \bottomrule
\end{tabular}%
}
\caption{Safety assessment results across model families. \xmark\ denotes \textit{unsafe} models, with a safety score below the threshold (i.e., $\tau = 0.5$), while \cmark\ indicates \textit{safe} models, with a score equal to or above the threshold.}
\label{tab:my-table}
\end{table}

\subsubsection{Responses to Research Questions}

We now summarize our findings by addressing the three research questions posed in Section~\ref{sec:intro}.

\vspace{0.5em}
\noindent \textbf{RQ1} \textit{How do different reasoning mechanisms (e.g., CoT prompting or reasoning by-design) affect robustness to bias elicitation?}

Our findings reveal that both forms of reasoning---whether elicited at inference time via CoT prompting or integrated by design in reasoning-enabled models---tend to amplify vulnerability to bias elicitation when compared to base models. Base models, which operate without explicit reasoning mechanisms, achieve the highest safety scores on average, indicating a stronger resistance to producing biased or harmful content. In contrast, the introduction of reasoning, regardless of the method, generally lowers safety performance. This suggests that reasoning as currently implemented may introduce additional pathways for stereotype reinforcement or rationalization. These results highlight a critical and somewhat counterintuitive insight, i.e., reasoning does not inherently improve robustness to bias and may, in fact, worsen it.

\vspace{0.5em}
\noindent \textbf{RQ2} \textit{Are reasoning models inherently safer than those relying on reasoning elicitation at inference time via CoT prompting?}

Our findings indicate that reasoning-enabled models are safer than those relying on reasoning elicitation through CoT prompting. On average, reasoning-enabled models outperform CoT-prompted variants in safety scores, showing lower rates of stereotypical responses. While both types of reasoning increase model complexity and may in general affect safety, CoT prompting appears more prone to generating harmful or biased content, likely due to its reliance on prompt-induced reasoning rather than internalized safety-aligned reasoning processes.

\vspace{0.5em}
\noindent \textbf{RQ3} \textit{How does the effectiveness of different jailbreak attacks targeting adversarial bias elicitation vary across reasoning mechanisms?}

Our findings highlight that model vulnerability is nuanced, varying with both the jailbreak strategy and the reasoning method used. 
CoT-prompted models are especially vulnerable to attacks involving low-resource languages or fictional storytelling that manipulate prompt context---framing it through reward incentives or role-playing scenarios---which can significantly affect models relying on prompt-induced reasoning paths not optimized for safety.
In contrast, reasoning-enabled models are more susceptible to obfuscation attacks like prefix injection or refusal suppression, which bypass internal safeguards by steering the model toward harmful outputs. This increased vulnerability likely stems from their greater generative freedom, enabling spurious justifications or rationalizations that align with the malicious instructions provided in the prompt.
Finally, base models tend to be the least vulnerable overall. Their simpler behavior and lack of explicit reasoning reduce the surface area for adversarial manipulation, making them comparatively more robust against a range of jailbreak strategies.

\section{Conclusion}
\label{sec:conclusion}

This study provides key insights into how different reasoning mechanisms affect robustness to bias elicitation in language models, using the CLEAR-Bias benchmark and the adversarial methodology proposed in~\cite{cantini2025benchmarking}. Our findings show that introducing reasoning—via inference-time CoT prompting or reasoning-enabled architectures—generally amplifies bias compared to non-reasoning base models. While reasoning-enabled models outperform those using zero-shot CoT prompting in safety, they still underperform base models overall. These results challenge the assumption that reasoning inherently aids bias mitigation and underscore the need for stronger safety alignment in reasoning-enabled language models.
There remain several avenues for future work. First, model behavior may vary with the formulation of CoT prompts, which can in turn affect safety. Second, reasoning traces can be analyzed to further understand how models justify responses to sensitive prompts. Emerging research suggests that models do not always ``\textit{say what they think}'', i.e., reasoning traces may not reflect internal decision-making processes~\cite{turpin2023language, chen2025reasoning}. Exploring these aspects can foster transparency and trustworthiness of reasoning language models, which is key in safety-critical applications.

\begin{acknowledgments}
We acknowledge financial support from “PNRR MUR project PE0000013-FAIR” - CUP H23C22000860006 and “National Centre for HPC, Big Data and Quantum Computing”, CN00000013 - CUP H23C22000360005. 
\end{acknowledgments}

\section*{Declaration on Generative AI}
  The author(s) have not employed any Generative AI tools.

\bibliography{bibliography}


\end{document}